\pdfoutput=1

\documentclass[11pt,a4paper]{article}
\usepackage{times,latexsym}
\usepackage{url}
\usepackage[T1]{fontenc}
\usepackage{bm}
\usepackage[dvipdfm]{graphicx} 
\usepackage{bmpsize}

\usepackage[acceptedWithA]{tacl2018v2}

\usepackage{hyperref}

\usepackage[utf8]{inputenc} % allow utf-8 input
\usepackage{booktabs}       % professional-quality tables
\usepackage{amsfonts}       % blackboard math symbols
\usepackage{nicefrac}       % compact symbols for 1/2, etc.
\usepackage{bbm}
\usepackage{calc}
\usepackage{enumitem}
\usepackage{booktabs}
\usepackage{multirow}
	
\usepackage{color, colortbl}

\usepackage{caption}
\usepackage{subcaption}

\usepackage{xspace}
\usepackage{amssymb}

\usepackage{amsmath}

\usepackage[normalem]{ulem} % temporary, just for strike-out with \sout.

\newcommand{\Table}[1]{Table~\ref{#1}}

\newcommand{\Figure}[1]{Fig.~\ref{#1}}
\newcommand{\Equation}[1]{Eq.~(\ref{#1})}

\newcommand{\Section}[1]{Sec.~\ref{#1}}
\newcommand{\ignore}[1]{}

\newcommand\ie{\emph{i.e.}}
\newcommand\eg{\emph{e.g.}}

\newcommand\vilbert{ViLBERT\xspace}
\newcommand\bert{BERT\xspace}
\newcommand{\modelname}[1]{MMT}

\DeclareRobustCommand{\mergedattn}{\emph{merged attention}\@\xspace} % concatenating image and language, called single stream before
\newcommand\coattn{\emph{coattention}\xspace} % cross query image and language, called multistream before
\newcommand\singleattn{\emph{modality-specific attention}\xspace} % independent self-attention pn image and language, called single modaliy stream before
\newcommand\langqattn{\emph{language-query attention}\xspace} % 
\newcommand\imageqattn{\emph{image-query attention}\xspace} % 

\newcommand\zsflickr{\emph{zero-shot Flickr}\xspace} % 
\newcommand\zsmscoco{\emph{zero-shot MSCOCO}\xspace} % 

\title{Decoupling the Role of Data, Attention, and Losses \\in Multimodal Transformers}

 \author{Lisa Anne Hendricks ~ John Mellor ~ Rosalia Schneider \\ {\bf Jean-Baptiste Alayrac} ~ {\bf Aida Nematzadeh} \\
 DeepMind
 \\ \tt \footnotesize{\{lmh, johnme, rgschneider, jalayrac, nematzadeh\} @google.com}}

\begin{document}

\maketitle

\begin{abstract}
Recently multimodal transformer models 
have gained popularity because their performance on language and vision tasks suggest they learn rich visual-linguistic representations. 
Focusing on zero-shot image retrieval tasks, we study three important factors which can impact the quality of learned representations: pretraining data, the attention mechanism, and loss functions.
By pretraining models on six datasets, we observe that dataset noise and language similarity to our downstream task are important indicators of model performance.
Through architectural analysis, we learn that models with a multimodal attention mechanism can outperform deeper models with modality-specific attention mechanisms.
Finally, we show that successful contrastive losses used in the self-supervised learning literature do not yield similar performance gains when used in multimodal transformers. \footnote{pre-print of MIT Press Publication version}
\end{abstract}

\section{Multimodal Pretraining}

Significant progress in pretraining of natural language processing (NLP) models has been made through both architectural innovations \citep[\eg, transformers;][]{vaswani2017attention} as well as a huge increase in the size of pretraining data and the model \citep[\eg,][]{devlin2018bert,brown2020language}.
This success in language pretraining has inspired parallel multimodal vision--language efforts; in particular, multimodal image--language transformers, pretrained on large noisy image--text datasets, have achieved state-of-the-art results on a range of downstream tasks such as image retrieval, visual question answering, and visual reasoning \citep[\eg,][]{lu2019vilbert,chen2019uniter,tan2019lxmert,Li2020unicoder,li2020oscar}. 

However, even though many variants of multimodal image--language transformer models have been proposed recently, it is unclear how learned representations are impacted by the large amounts of pretraining data, the transformer architecture and self-attention, or their specific losses. We address this gap, by first establishing a baseline that is trained on the same pretraining data as multimodal transformers but with a different architecture. We then perform an investigative analysis to better understand the extent to which these aspects contribute to models' performance.

Our evaluation mainly focuses on zero-shot tasks where evaluation data is taken from a dataset unseen during pretraining. Measuring zero-shot performance enables us to evaluate whether a pretrained model learns general representations.
Previous work in NLP has considered probing classifiers to evaluate representations; however, this approach can be misleading as the performance of probing classifiers does not solely depend on the quality of representations \citep[\eg,][]{hewitt-liang-2019-designing,voita2020information}. Similarly, evaluation after fine-tuning is a less direct measure of strength of representations since performance on these tasks is highly dependent on the fine-tuning experimental set-up and the size of fine-tuning data \citep{yogatama2019learning}.

We first study the importance of different properties of multimodal datasets such as their size and their noise level (\ie, how closely the language describes a given image's content). Recent work has introduced image--text datasets with different qualities, for example, noisy but very large ones \citep{sharma2018conceptual} as well as carefully-annotated but smaller ones \citep{pont2019connecting}. Better understanding of what aspect of a dataset is more important can result in better task performance and also guide us in future dataset curation efforts. We find that a dataset's size does not always predict multimodal transformers' performance; its noise level and language similarity to the evaluation task are both important contributing factors. We also show that multimodal transformers can achieve competitive results without relying on language-only or image-only pretraining for weight initialization or feature extraction.

We also dissect multimodal transformers' architecture, analyzing the effectiveness of different attention mechanisms, depth, and number of parameters. We show that \emph{multimodal} attention where both language and image % modalities
transformers attend to each other are crucial for these models' success. Multimodal attention achieves the best results when combined with multi-level (deep) interactions. Moreover, models with other types of attention (even with more depth or parameters) fail to achieve comparable results to shallower and smaller models with multimodal attention.

Additionally, inspired by the success of \citep[\eg,][]{oord2018representation} for self-supervised representation learning, we examine whether using a contrastive image--text matching loss instead of a classification one improves the quality of representations in our models. Surprisingly, we find that the choice of image--text matching loss does not matter much in multimodal transformers. %(with multimodal attention); 
On the other hand, models \emph{without} multimodal attention (a multi-level ``cross-talk'' between modalities) benefit significantly from a contrastive loss.

Finally, we believe that advances in multimodal pretraining can have significant impacts on a wide range of downstream applications; however, it is important to form a clear understanding of how and why multimodal transformer models perform well to avoid overfitting to a set of downstream evaluation tasks.
Our analysis of pretraining data, attention, and loss functions is an important step towards gaining a deeper understanding of these powerful models.

\section{Multimodal Transformers}
The success of transformer-based language models on a variety of language tasks~\citep[\eg,][]{devlin2018bert} has inspired similar multimodal efforts~\citep[\eg,][]{lu2019vilbert,chen2019uniter,tan2019lxmert,Li2020unicoder,li2020oscar}.\footnote{We use the term multimodal transformers to refer to image--text transformer--based models. Note that similar architectures are applied to other modalities such as videos \citep{sun2019videobert} but are outside of the scope of this work.}
The main distinction is that image-text multimodal transformers take image-text pairs as input, attend over both modalities, and are trained with additional losses. Similar to the language models, multimodal transformers are often fine-tuned on down-stream tasks but multimodal ones; \eg, image retrieval \citep{young2014image} or visual question answering \citep{goyal2017making}.

We give a brief overview of the \bert model~\citep{devlin2018bert} which forms the backbone of multimodal transformers. The \bert architecture consists of a stack of transformer blocks \citep{vaswani2017attention} and has three main components. 
First, the input text is tokenized and three embedding functions are used to embed the token, its position in the sentence (\ie, positional encoding), and the sentence it belongs to. 
The final {language} embedding is a summation of these three vectors. 
The BERT model also includes a \texttt{<SEP>} token to separate different sentences and a \texttt{<CLS>} token which can be thought of as an aggregate representation of the input text.
Second, the sequence of token embeddings are input into a series of transformer layers where tokens are combined through self-attention. Third, two different losses are applied to the model output: a \emph{masked language modeling} loss, in which the model predicts a masked word (denoted by a \texttt{<MASK>} token), and a \emph{next sentence prediction} loss which, given two sentences, predicts if the second sentence follows the first.

Multimodal transformer models facilitate learning from multimodal data via three changes to the \bert architecture: \emph{multimodal data preprocessing} (more specifically images), adding \emph{multimodal attention} by changing self-attention such that it combines image and text modalities, and introducing image and multimodal \emph{loss functions}.

\subsection{Multimodal Data Processing} 
\begin{figure*}
\centering
  \caption{ Tracking queries, keys and values for different attention types described in \protect\Section{sec:mmattention}.  }

  \includegraphics[width=\linewidth]{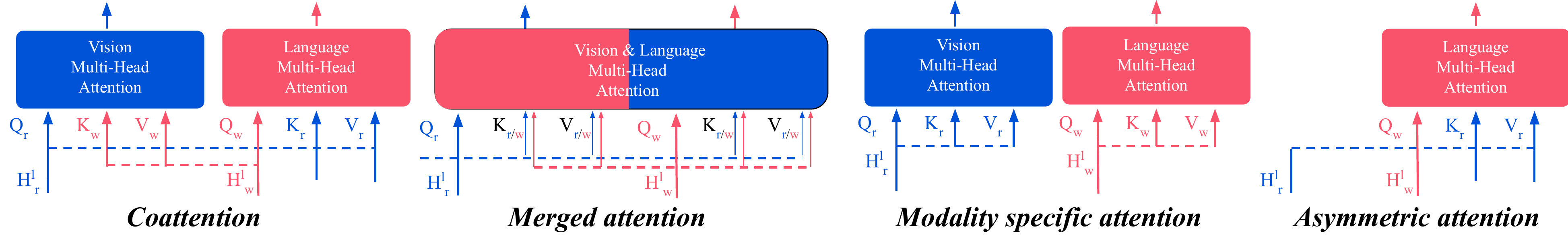}

  \label{fig:attention}
  \vspace{-0.4cm}
\end{figure*}

Training multimodal transformers requires image--text pairs such that the text for a given image, at least to some degree, describes the image. 
Recent work attempts to remove the annotation cost by automatically collecting datasets \citep[\eg, web images and their alt-text as in][]{sharma2018conceptual}. In \Section{sec:data_results}, we examine whether the quality of text descriptions impacts these models' performance.

The text input processing is the same as language models; in fact, many of the existing models \cite[such as][]{lu2019vilbert} are initialized with \bert pretrained weights. We show that this initialization is not important in our experiments (see \Section{sec:data_results}).
Processing images into a sequence involves defining ``visual tokens'' analogously to language tokens. Almost all image-text multimodal transformer models consider a bounding box from a pretrained object detection model to be a ``visual token''. 
Similar to the positional encodings in language models, for each visual token, the spatial position of each bounding box is also encoded.

Although most multimodal transformers require training a supervised model (a  detector) to extract bounding-box features, there are other possible ways to represent visual tokens -- for example, \citet{huang2020pixel} bypass training a detector by using regions from a high-level feature map in an image classification network as visual tokens.
We focus our studies on models which use bounding-box features as this reflects the majority of recent work, though we achieve comparable results when learning directly from images without a detector (or even a pretrained classifier) in \Section{sec:data_results}.

\subsection{Multimodal Attention}
\label{sec:mmattention}

Each transformer block consists of a multi-head attention module \citep{vaswani2017attention} that for a given token embedding produces a weighted representation of all other 
tokens in a sentence. 
This weighted representation is then combined with the input representation of the given token and is passed to the next layer. More specifically, for the token $i$ at layer $l$, each attention head takes as input a key $k^i_l$, value $v^i_l$, and query $q^i_l$ which are computed by passing the representation from the previous layer $h^i_{l-1}$ through a linear layer. The output of the attention module for token $i$ is:
\begin{equation}
    A(q^i_{l}, K_l, V_l) = \text{softmax}\left(\frac{q^i_{l}K_{l}}{\sqrt{d_k}}\right) V_{l}
\end{equation}
where $d_k$ is the dimension of the key and $K_l$ and $V_l$ matrices contain all tokens' keys and values. 

Given this definition, there are a few possible ways to implement multi-head attention over image and language modalities as shown in \Figure{fig:attention}.
For a given query (from one modality), we can simply consider keys and values from all input tokens regardless of the modality type \cite[\eg,][]{chen2019uniter}. 
We refer to this multimodal attention as \mergedattn since it simply merges inputs from the two modalities.

Alternatively, given queries from one modality (\eg, image), keys and values can be taken \emph{only} from the other modality (\eg, language). Following \citet{lu2019vilbert}, we refer to this multimodal attention as \coattn. 
We also consider cases where this attention is asymmetric, \ie, queries are \emph{either} from language or image, while keys and values are from image or language, respectively. We call these two attention types \langqattn or \imageqattn.

Another possibility is to consider single-modality transformers where queries, keys, and values all come from either the image or text modality; we refer to this attention as \singleattn where each modality has its own multi-head attention. Single-modality transformers with modality-specific attention allow us to study the role of ``cross-talk'' between modalities in multimodal transformer models. 

We note that we use the term \emph{multimodal} attention to refer to both \mergedattn and \coattn and discuss the importance of different attention types in \Section{sec:attention_results}.

\subsection{Multimodal Loss Functions}
\label{sec:loss}

Broadly, multimodal transformers have three loss types, language and image losses that are applied to the language and image outputs, respectively, as well as an image-text matching loss applied to image--language pairs.
Let $\bm{r}=\{r_1, \cdots, r_N\}$ be the $N$ input image regions and $\bm{w}=\{w_1, \cdots, w_T\}$ be the $T$ word tokens representing an image--text pair.
A subset of input image regions and word tokens are masked (\eg, set to zero) before being passed through the transformer layers. After applying the mask, we refer to the unmasked image regions as $\bm{r}^{\bm{m}}$ and to the unmasked word tokens as $\bm{w}^{\bm{m}}$. We use $N_m$ and $T_m$ to denote the set of image region and word token indices that are masked, respectively.
Similar to the BERT model, the language loss is a masked-language modelling (MLM) loss:
\begin{equation}
  -\sum_{t \in T_m}\log P^{\bm{w}}_\theta(w_t | \bm{w}^{\bm{m}}, \bm{r}^{\bm{m}}),
  \label{eq:mlm}
\end{equation}
where $P^{\bm{w}}_\theta$ corresponds to the output probability distribution over words in the vocabulary from the transformer model parameterized by $\theta$.

Most models also include an analogous \emph{masked region modeling loss} (MRM) for images.
One popular region modelling loss, for each bounding box, minimizes the KL-divergence between the predicted distribution over object classes and the distribution over classes obtained from a pretrained detector $D(l|r_n)$ \cite[\eg,][]{chen2019uniter, lu2019vilbert}.
\begin{equation}
     \sum_{n \in N_m} \mathrm{KL}(D(l|r_n) || P^{\bm{r}}_\theta(r_n | \bm{r}^{\bm{m}}, \bm{w}^{\bm{m}})),
    \label{eq:mrm}
\end{equation}
where 
$P^{\bm{r}}_\theta$ corresponds to the predicted probability distribution over object classes from the transformer model parameterized by $\theta$.

Finally, multimodal transformer models include an image--text matching (ITM) loss which predicts whether an image and text pair match; this is generally posed as a binary classification problem:
\begin{align}
 &-y\log(\sigma(s_{\theta}(\bm{r}^{\bm{m}}, \bm{w}^{\bm{m}}))) \nonumber \\
 -&(1-y)\log(1-\sigma(s_\theta(\bm{r}^{\bm{m}}, \bm{w}^{\bm{m}}))), 
 \label{eq:itm-cls}
\end{align}
where $y$ is equal to $1$ for positive pairs and $0$ otherwise and $s_{\theta}$ corresponds to the confidence score of the model that a pair $(\bm{r}, \bm{w})$ are matched and $\sigma$ is the sigmoid function.
Recently, contrastive image--text matching losses have been successful in self-supervised representation learning \citep[\eg,][]{oord2018representation}; thus, we also explore whether a contrastive formulation of ITM can improve the performance of multimodal transformers and discuss the challenges of using these losses for multimodal transformer models. Our contrastive loss is formulated as:
\begin{equation}
 -\log\left(\frac{e^{s_\theta(\bm{r}^{\bm{m}},\bm{w}^{\bm{m}}))}}{e^{s_\theta(\bm{r}^{\bm{m}}, \bm{w}^{\bm{m}})}+\sum\limits_{(\tilde{\bm{r}},\tilde{\bm{w}})\sim\mathcal{N}}\hspace*{-4mm}e^{ s_\theta(\tilde{\bm{r}}^{\bm{m}}, \tilde{\bm{w}}^{\bm{m}})}}\right),
 \label{eq:itm-contrastive}
\end{equation}
where $\mathcal{N}$ is a set of negative image-text pairs.
\Section{sec:loss_results} outlines our findings on loss ablations.

\section{Experimental Setup}

Here we outline the details of our experimental setup: the base multimodal transformer model used in most of our experiments, our baseline model, and the pretraining datasets.

\subsection{Base Multimodal Transformer}

Our base multimodal transformer model (\modelname{}) most closely resembles the \vilbert model \citep{lu2019vilbert}. For text inputs, we first tokenize sentences using SentencePiece \citep{kudo2018sentencepiece} and truncate sentences into a fixed length of $22$ for pretraining datasets and $25$ for datasets used to fine-tune and evaluate retrieval models. 
We  then include a separator (\texttt{<SEP>}) and an aggregator (\texttt{<CLS>}) token. % for a total length of $22$. 
Unless otherwise stated, we do not transfer weights from a pretrained BERT model.

For image inputs, we represent ``visual tokens'' as region of interest (RoI) pooled features corresponding to bounding boxes from an object detector \cite{ren2015faster} trained on Visual Genome \cite{krishna2017visual} images with labels parsed as was done in \citet{anderson2018bottom}. 
The detection model is trained using a multi-label sigmoid cross-entropy loss to simultaneously predict objects and attributes. 
The highest $36$ or $100$ scoring bounding boxes are input when pretraining or evaluating, respectively. 
Like \vilbert, we include an ``average'' feature which is computed by averaging features across bounding boxes and serves a similar role to the \texttt{<CLS>} token in the text input. 

In addition to the positional encoding added to text embeddings before the first transformer layer, we also add the positional encoding to the text embedding \emph{at each layer} of the language-only transformer blocks as in XLNet \cite{yang2019xlnet} because this led to improvements on a language--only BERT model. 
For image inputs,
we embed bounding box coordinates and add this to our image embedding.

In our model, following \vilbert, a multimodal co-attention layer consists of an image-only and a language-only transformer, each followed by a transformer with \coattn (see \Section{sec:mmattention}). We use the term ``layer'' to refer to this multimodal layer. Like VilBERT, our model consists of $6$ language-only layers, followed by $6$ multimodal ones.
We train the model by minimizing masked language modelling (\Equation{eq:mlm}), masked region modeling (\Equation{eq:mrm}), and binary classification image--text matching (\Equation{eq:itm-cls}) losses.
To calculate the image-text loss, we apply an element-wise multiplication to the \texttt{<CLS>} language features and output corresponding to the averaged image feature input.
The resulting ``multimodal feature'' is input into a classification model.  
We create negative image-text examples by sampling text from another image in our batch. Unless otherwise noted, we have an equal number of negative and positive image-text pairs.

We train our models with a global batch size of 1024 distributed over 64 Google Cloud TPU v3 cores\footnote{https://cloud.google.com/tpu/}.
We use the LAMB optimizer \citep{you2019large} with an initial learning rate of $0.00176$ and 20,000 warm up steps.
Learning rate is decayed with polynomial decay with a minimum learning rate ratio of $0.004$.
We use gradient clipping ($1$) and dropout ($0.1$) as well as weight decay ($0.1$). We find weight decay particularly important in ensuring that our loss did not diverge.
We train our models for a maximum of 1,000,000 iterations.

\subsection{The Baseline Model}
\label{sec:baseline_model}

Multimodal transformers are different from most prior image--text models because they are pretrained on a large dataset (millions of image-text pairs).
To better understand if data alone can lead to better image--text representations, we train a strong baseline model, which does not include a multimodal attention mechanism, with the same data as our multimodal transformer.
% To investigate how these differences have an impact on multimodal transformers, we implement a baseline model without multimodal attention, but pre-trained on the same data as our multimodal transformer.

Our baseline model learns a joint space between language and vision~\citep{weston11wsabie,frome13devise,kiros2014unifying} by minimizing the distance between image and text features taken from a positive pair (where text describes the image) and at the same time increasing that distance for a negative pair. 
Despite lacking a multimodal attention mechanism, this approach has been popular in image and video domains due to its simplicity and effectiveness for retrieval applications~\citep[\eg,][]{gong14multi,wang2016learning,chowdhury2018webly,miech18learning}.

To implement our baseline, we first encode word tokens $\bm{w}$ into a fixed-size sentence representation $S \in \mathbb{R}^{768}$ and image regions $\bm{r}$ into a fixed-size image representation $I \in \mathbb{R}^{768}$.  
To encode sentence representations, we input words into a randomly initialized BERT model and extract sentence representations $S$ from the \texttt{<CLS>} output.
To extract image representations $I$, we first mean-pool features across detected bounding boxes
and then pass the mean-pooled features into a one-layer MLP with an output of size $768$.
Finally, we element-wise multiply $I$ and $S$ and input the resulting vector into a two-layer MLP parameterized by $\theta$ which outputs a score, $s_{\theta}$ indicating whether $I$ and $S$ match.
We train our baseline model using the contrastive loss defined in Equation~\eqref{eq:itm-contrastive} with 1024 negative examples.
The detector weights are fixed during training.

\subsection{Pretraining Datasets}
\label{sec:pretraining_data}

\begin{table}[]
\caption{\small The pretraining datasets: the type and number of images and captions.}
\resizebox{\linewidth}{!}{
\begin{tabular}{@{}lrlr@{}}
\toprule
\multirow{2}{*}{Dataset}                 & \multirow{2}{*}{\# images}  & \multicolumn{2}{c}{Caption} \\
                                         &                             & Type           & \#         \\ \midrule
\multicolumn{1}{l|}{MSCOCO}              & \multicolumn{1}{r|}{83K} & Annot.         & 592K     \\
\multicolumn{1}{l|}{Visual Genome (VG)}  & \multicolumn{1}{r|}{110K}   & Annot.         & 5.4M       \\
\multicolumn{1}{l|}{MSCOCO-narratives}   & \multicolumn{1}{r|}{83K}    & Narration      & 230K       \\
\multicolumn{1}{l|}{OI-narratives}       & \multicolumn{1}{r|}{500K}   & Narration      & 1.3M     \\
\multicolumn{1}{l|}{SBU}                 & \multicolumn{1}{r|}{1M}     & Web            & 1M         \\
\multicolumn{1}{l|}{Conceptual Captions} & \multicolumn{1}{r|}{2.7M}   & Alt-text       & 2.7M       \\ \bottomrule
\end{tabular}
}
% \vspace{-.4cm}
\label{tb:pretraining_stats}
\end{table}

\emph{Conceptual Captions (CC)} consists of over 3 million image-text pairs harvested from the web where the caption corresponding to an image is its alt-text description \citep{sharma2018conceptual}.
Image--text pairs are filtered and preprocessed such that text is more image relevant than raw Alt-text; 
however, the dataset is still ``noisy'' and includes pairs where the text is not relevant to the image's content. 
We were able to download 81\% of the training set 
of CC; unless otherwise stated, we train our models on this subset of CC.

The \emph{SBU} dataset \citep{ordonez2011im2text} consists of 1 million image-text pairs sourced from Flickr with text taken from users' captions.
As a result, similar to CC, not all text is image relevant. 
We also use datasets which were collected by asking annotators to describe images, resulting in more image relevant language including the \emph{MSCOCO} %image-caption 
dataset \citep{chen2015microsoft} and \emph{Visual Genome (VG)} \citep{krishna2017visual}, which
includes descriptions for bounding boxes in images.
When using VG, we consider each bounding box description to be a caption for the entire image.

We also experiment with the \emph{Localized Narratives} dataset \citep{pont2019connecting}. 
This dataset includes rich annotations collected by asking users to describe an image while pointing to each part of the image being described (using their mouse). The resulting ``narratives'' often consist of multiple sentences.
We break the narratives into individual sentences and treat each sentence as a caption paired with the image. We use the localized narratives collected for the Open Images \citep{kuznetsova2018open} and MSCOCO datasets, and refer to them as OI-narratives and MSCOCO-narratives. This allows us to compare models which are trained with the same images (MSCOCO) with different language (MSCOCO captions vs. localized narratives).
\Table{tb:pretraining_stats} provides an overview of our pretraining datasets.

Finally, we consider combining datasets using two sampling approaches: \emph{instance sampling} where we mix all datasets together and sample from this mix for each batch and \emph{dataset sampling} where we sample evenly from datasets so that each batch contains the same number of examples from each dataset.
For datasets with multiple captions, we first sample an image, then sample a caption for the given image.
We combine all six datasets described here as well as the four datasets combined in \citet{chen2019uniter} (MSCOCO, VG, SBU, and Conceptual Captions) which we refer to as UNITER data.

\begin{table}[]
\caption{\small Number of images in evaluation tasks and whether datasets were used in a zero-shot (ZS) or fine-tuned (FT) setting.}
\centering
%\resizebox{\linewidth}{!}
{
\begin{tabular}{@{}lrrcc@{}}
\toprule
\multirow{2}{*}{Dataset}                 & \multicolumn{2}{c}{\# images}  & \multirow{2}{*}{ZS} & \multirow{2}{*}{FT} \\
& train & test & \\
 \midrule
\multicolumn{1}{l}{Flickr30k}              & \phantom{0}29K & \phantom{00}1K   & \checkmark  & \checkmark\\
\multicolumn{1}{l}{MSCOCO}  & n/a   &  \phantom{00}5K  & \checkmark &       \\
\multicolumn{1}{l}{VQA}   &  440K   & 210K  & & \checkmark       \\
\bottomrule
\end{tabular}
}
% \vspace{-.4cm}
\label{tb:ft_datasets_stats}
\end{table}

\subsection{Evaluation Tasks}

We focus on \emph{zero-shot} evaluation as it enables us to examine the representations without confounding our findings with the side-effects of fine-tuning \citep[][]{yogatama2019learning} or probing classifiers \citep[\eg,][]{zhang2018language,hewitt-liang-2019-designing}. Following \citet{lu2019vilbert} and \citet{chen2019uniter}, we use the term \emph{zero-shot} to refer to experiments where we test our models on a dataset different from our pretraining data \emph{without fine-tuning}. For example, we use the MSCOCO dataset to test the models that are pretrained on Conceptual Captions. This is considered as a zero-shot task since the properties of the dataset used for testing (for example, its language) differ from those in the pretraining dataset.
We use zero-shot image retrieval tasks since image retrieval directly measures what our pretraining data and objectives encourage our models to learn: whether an image and a sentence are aligned.

We evaluate on the Flickr30k dataset \cite{young2014image} (referred to as \zsflickr) and use the splits defined in \citet{karpathy2015deep}.
We evaluate checkpoints after 1 million steps as well as when the loss on the CC validation set is lowest.
When varying the pretraining data, our models sometimes overfit quickly on smaller datasets; as a result, we evaluate checkpoints every 100K steps.
We select the best checkpoint according to zero-shot performance on Flickr30k val and use it for all other downstream tasks.
We also report retrieval numbers on MSCOCO \citep{chen2015microsoft} (which we call \zsmscoco) using the splits of \citet{karpathy2015deep}. 
Reported retrieval numbers are on the test split of datasets.
Images in Flickr30k and MSCOCO are annotated with $5$ captions.

In addition to the zero-shot image retrieval tasks, we use the fine-tuned Flickr30k image-retrieval task to examine whether our observations transfer when fine-tuning the \modelname{} model.
We fine-tune our models for 10,000 steps and use MLM, MRM, and ITM losses.
All results for image retrieval are reported using Recall@K (R@K), which measures whether the ground-truth image is among the top K images retrieved by our  model.

When comparing pretraining datasets, we hypothesize that which pretraining dataset is best depends on the downstream task, so we additionally consider VQA \cite{antol2015vqa,goyal2017making}. 
To fine-tune for VQA, 
we replace the image--text matching loss with a 2-layer MLP and train with a binary cross-entropy loss against soft answer scores \cite{teney2018tips}.
We use similar hyper-parameters as when pretraining 
and report results on the validation set.
We report the average score across 3 random initializations of the MLP.

We use Flickr IDs to filter out images appearing in the Flickr30k and MSCOCO validation/test sets from our pretraining sets.
Conceptual Captions is not collected from Flickr, so we could not filter out images using this method.
Table \ref{tb:ft_datasets_stats} provides an overview of our evaluation datasets.

\section{Experimental Results}
We first compare \modelname{} to a baseline and then investigate how pretraining data, attention, and loss functions impact model performance.

\subsection{Comparison to a Baseline}

We compare our multimodal transformer (\modelname{}) against a strong 
baseline  inspired by recent success in visual retrieval \citep[\eg,][]{miech18learning}.
To disentangle the effect of pretraining data and architecture, we investigate whether our baseline (described in \Section{sec:baseline_model}), without \emph{multimodal} attention or MLM and MRM losses but pretrained on the same data (\ie, Conceptual Captions) as multimodal transformers produces competitive results. 

In Table \ref{tab:baseline}, we compare \modelname{} to our proposed baseline, verifying that \modelname{} learns better representations not only because it is pretrained on a large dataset, but because of architectural choices. 
Our \modelname{} results are on par with existing models trained with the same 
data: comparing to \vilbert, the most similar model to ours, on the \zsflickr, we achieve an R@1 of $41.9$ in comparison to $31.9$.
As expected, retrieval numbers on \zsmscoco are lower than \zsflickr because MSCOCO has more images in its evaluation set (see Table \ref{tb:ft_datasets_stats}) and is therefore harder.
On the fine-tuned image retrieval task, we achieve comparable performance to \vilbert (our R@1 is $59.1$ vs. $58.2$), even though we do not sample hard negatives when training.
We emphasize that our goal is not to outperform existing work, but to build a strong multimodal transformer model to analyze the role of data, attention, and losses.

We verify that a contrastive loss (\Equation{eq:itm-contrastive}) leads to stronger results than a classification one. 
As shown in Table \ref{tab:baseline}, replacing the contrastive loss with a classification loss consistently decreases performance.
Initializing our baseline with BERT weights marginally decreases performance, \eg, R@1 on \zsflickr decreases by $0.6$.

\begin{table}[t]
 % \vspace{-0.2cm}
\caption{\small Comparison of our proposed baseline to our multimodal transformer model (\modelname{}).}
\centering
 % \vspace{-0.15cm}
\resizebox{\linewidth}{!}{
\begin{tabular}{@{}l|cc|cc|cc|@{}}
\toprule
                              & \multicolumn{4}{c|}{Flickr30k}                                       & \multicolumn{2}{c|}{MSCOCO} \\
                              & \multicolumn{2}{c}{ZS}           & \multicolumn{2}{c|}{FT}           & \multicolumn{2}{c|}{ZS}     \\
\multicolumn{1}{l|}{}         & R1  & \multicolumn{1}{c|}{R10} & R1  & \multicolumn{1}{c|}{R10} & R1          & \multicolumn{1}{c|}{R10}        \\ \midrule
 Baseline & 25.4 & \multicolumn{1}{c|}{64.9} & 40.9 & \multicolumn{1}{c|}{81.8} & 13.0         & \multicolumn{1}{c|}{44.5}        \\
{$-$ contrastive} & {21.7} & \multicolumn{1}{c|}{61.0} & {39.0} & \multicolumn{1}{c|}{80.6} & {10.2}  & \multicolumn{1}{c|}{40.9}        \\
{$+$ BERT PT} & {24.8} & \multicolumn{1}{c|}{65.1} & {39.9} & \multicolumn{1}{c|}{79.9} & {12.7}         & \multicolumn{1}{c|}{43.1}        \\
\midrule
\modelname{}     & \textbf{41.9} & \multicolumn{1}{c|}{\textbf{79.0}} & \textbf{59.1}  &  \multicolumn{1}{c|}{\textbf{91.5}} & \textbf{21.3}        & \multicolumn{1}{c|}{\textbf{57.9}}        \\
%\midrule
\vilbert & 31.9 & \multicolumn{1}{c|}{72.8} & 58.2 & \multicolumn{1}{c|}{\textbf{91.5}} & - & \multicolumn{1}{c|}{-} \\
%\\ \bottomrule
\end{tabular}}
 \label{tab:baseline}
 % \vspace{-0.45cm}
\end{table}

\subsection{Multimodal Data Preprocessing}
\label{sec:data_results}
We investigate how pretraining datasets, supervised image features, and weights from a pretrained language model impact our results.

\begin{figure}[t!]
        \centering
        \hfill
        \caption{\small Effect of pretraining data. The datasets on X axis are ordered based on their \zsflickr scores.  IS:Instance Sampling, DS: Dataset Sampling.}
        \begin{subfigure}[]{.5\textwidth}
            \centering
            \caption{\small Zero-shot (ZS) \& fine-tuned (FT) image retrieval (IR) }
            \includegraphics[width=.9\linewidth]{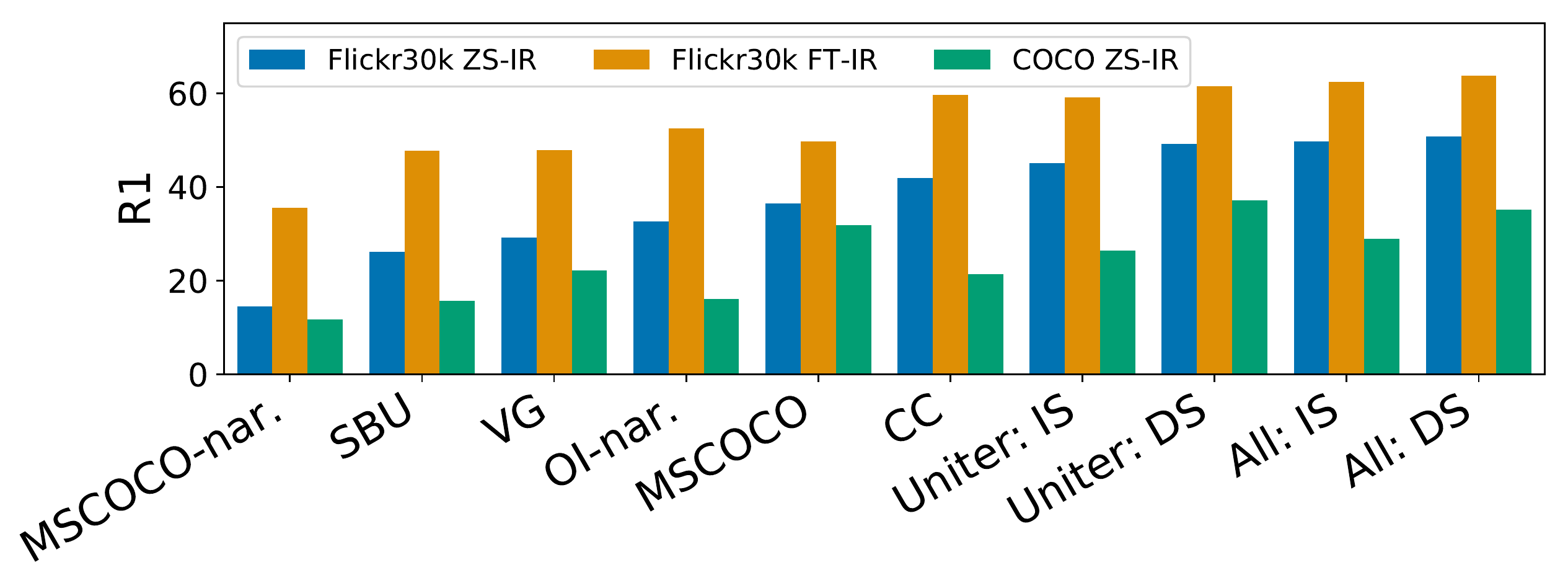} 
            \label{fig:pre-data-ir}
        \end{subfigure}
        \begin{subfigure}[]{.5\textwidth}
            \centering
            \caption{\small Visual question answering (VQA v2)}
            \includegraphics[width=.9\linewidth]{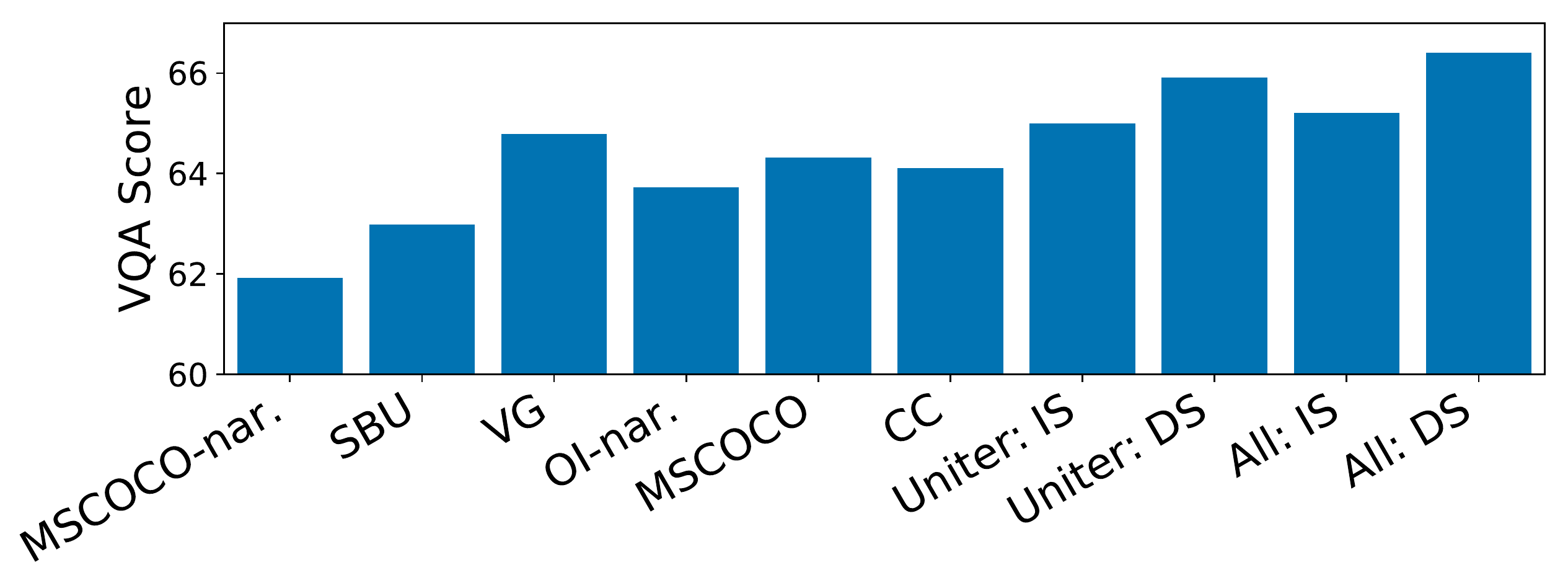} 
            \label{fig:pre-data-vqa}
        \end{subfigure}
        % \vspace{-0.1cm}
        
        % \vspace{-0.6cm}
        \label{fig:pre-data}
\end{figure}

\paragraph{Pretraining Datasets.} \Figure{fig:pre-data} reports our results when we pretrain the \modelname{} on the individual and combined datasets introduced in \Section{sec:pretraining_data}. 
We observe that in all our tasks, \textit{larger datasets usually lead to better performance, but not always.} For example, SBU consistently performs worse than MSCOCO, despite being substantially larger. 

Additionally, \emph{when combining datasets, how datasets are sampled matters}.
In our experiments, dataset sampling (DS) is more effective than instance sampling (IS).
In dataset sampling, smaller datasets (like MSCOCO) will be sampled more frequently than in instance sampling. 
Since MSCOCO pretraining leads to good performance, more exposure to MSCOCO samples is beneficial.
We consider combining all datasets as well as 
datasets combined in UNITER
\cite{chen2019uniter}.
\Figure{fig:pre-data-ir} shows that combining all datasets performs better than UNITER data on the \zsflickr task, but not on the \zsmscoco, showing that more data is not always better.
On \zsmscoco the impact of the sampling mechanism is even more evident: given UNITER data, dataset sampling performs better than instance sampling by over $10$ points ($37.1$ vs $26.4$).

Next, we compare datasets that have a similar number of images to investigate the role of the type of language used in each dataset.
As an extreme example, MSCOCO and MSCOCO-narratives contain the same images, but the former does substantially better on our downstream tasks.
To better understand this observation, we quantify the difference between the language of pretraining and evaluation datasets: we trained a language model (a 6-layer Transformer) on a given pretraining dataset, and use that model to compute the perplexity of the evaluation dataset. 
For our three datasets with the same number of images (MSCOCO, MSCOCO-narratives, and VG), the perplexity of the evaluation dataset (Flickr or MSCOCO) explains their performance -- the perplexities are the lowest on MSCOCO, then VG, and lastly on MSCOCO-narratives. 
\emph{This shows that the similarity between the language of pretraining and evaluation datasets is important.}

However, not all performance differences are explained by the number of images or perplexity: pretraining on SBU results in poorer performance than OI-narratives on our downstream tasks, despite SBU having twice the number of images and lower perplexity on both evaluation datasets.
We conjecture that SBU's poor performance is due to noise: SBU text is scraped from captions and may not match the images as well as the manually annotated text in OI-narratives.
To investigate this, we calculate an \emph{overlap metric} for an image--text pair as the ratio of text words overlapping with predicted bounding box labels. 
For each dataset, we calculate the average overlap for $3000$ images, providing an approximation of how much the language describes the images in the dataset.
The overlap is much lower for SBU compared to OI-narratives (0.14 vs. 0.25), showing that SBU is indeed noisier, which can decrease its utility for pretraining multimodal representations.\footnote{The \emph{overlap metric} for other datasets: VG: 0.82, MSCOCO: 0.42, MSCOCO-narratives: 0.27, and CC: 0.11.}

Moreover, we observe that the \textit{goodness of a pretraining dataset for one task does not always transfer to a different task.} 
For example, CC is a better pretraining dataset than VG when fine-tuning for image retrieval, but they perform similarly when fine-tuning for VQA, a substantially different task.
In fact, we note that VQA performance varies less across pretraining datasets (\eg, CC, VG, and MSCOCO), likely because the VQA training split is large.
We also observe differences between zero-shot and fine-tuned image retrieval. 
Though MSCOCO performs $3.8$ points better on \zsflickr than OI-narratives, OI-narratives performs $2.9$ points better after fine-tuning.

\begin{figure}[t!]
        \centering
          \caption{\small Comparing models trained with the MSCOCO and CC datasets. We provide the top-1 ranked retrieved image given an input query sentence on the Flickr val dataset. Correctly retrieved images are framed in green and the incorrect ones in red.}
        % \vspace{-.2cm}
        \includegraphics[width=\linewidth]{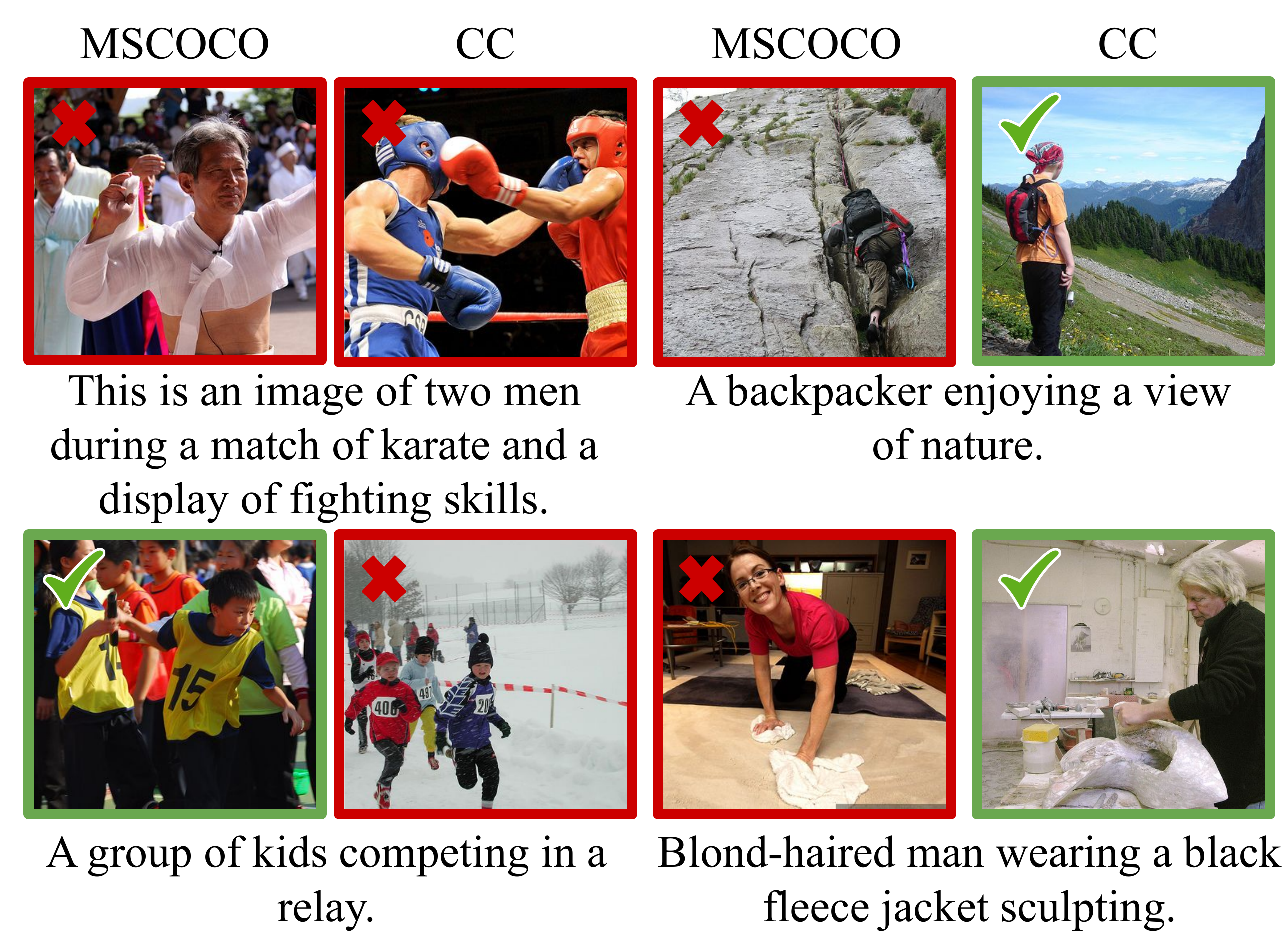}
        \vspace{-0.6cm}
        \label{fig:qual-coco}
\end{figure}

Finally, to visually illustrate the difference between the learned representations, we compare qualitative examples of models trained with our best two pre-training datasets: MSCOCO and CC (see \Figure{fig:qual-coco}).
Though the model trained with MSCOCO retrieves examples with some semantic relevance, our model trained with CC is able to retrieve images with more correct details like ``enjoying a view'' and ``black fleece jacket''.

\begin{figure*}[t!]
        \centering
        \caption{\label{fig:ablation_graphs} Ablation studies on number of layers and heads.
        }
        \vspace{-.4cm}
        % \hfill
        
        \begin{subfigure}[t]{.3\linewidth}
            \centering
                \includegraphics[height=2.5cm]{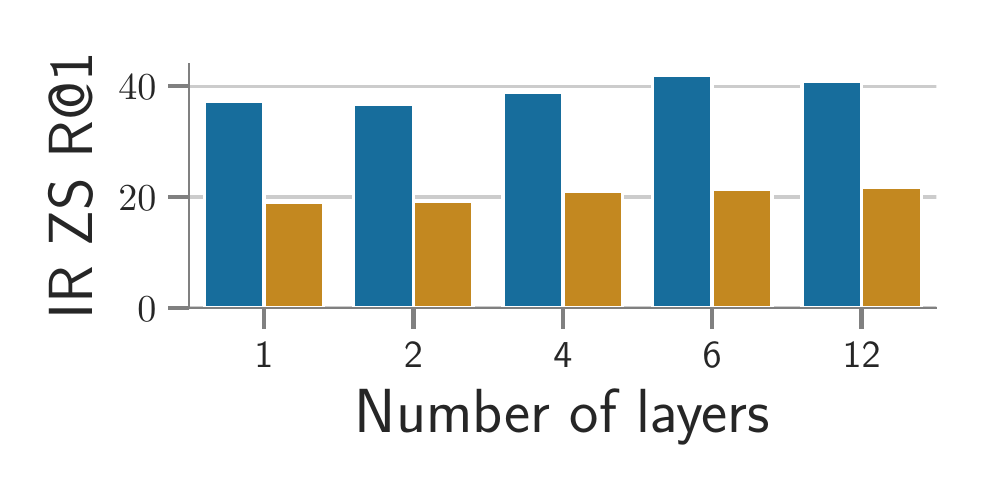} % second figure itself
                \caption{\small Depth \label{fig:abl_layers}}
        \end{subfigure}
        % \hfill
        \begin{subfigure}[t]{.3\linewidth}
            \centering
                \includegraphics[height=2.5cm]{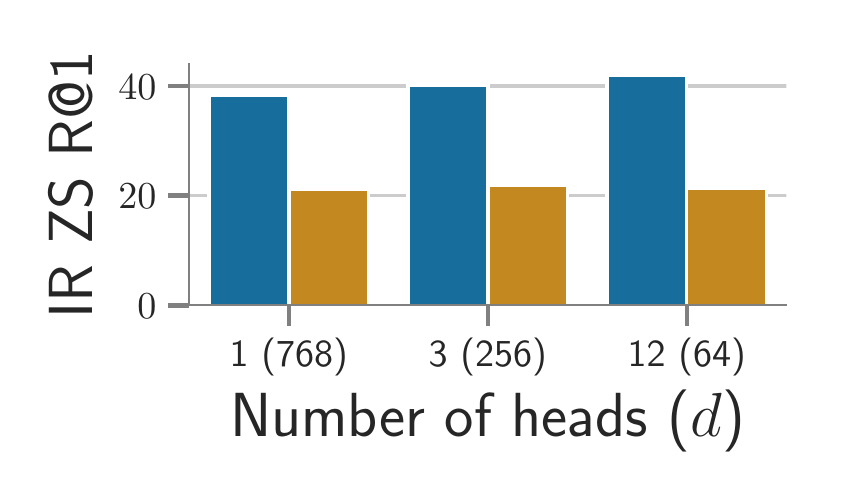} % first figure itself
                \caption{\small Heads: fixed \#params \label{fig:abl_heads_params}}
        \end{subfigure}
        % \hfill
        \begin{subfigure}[t]{.3\linewidth}
                \centering
                % TODO
                \includegraphics[height=2.5cm]{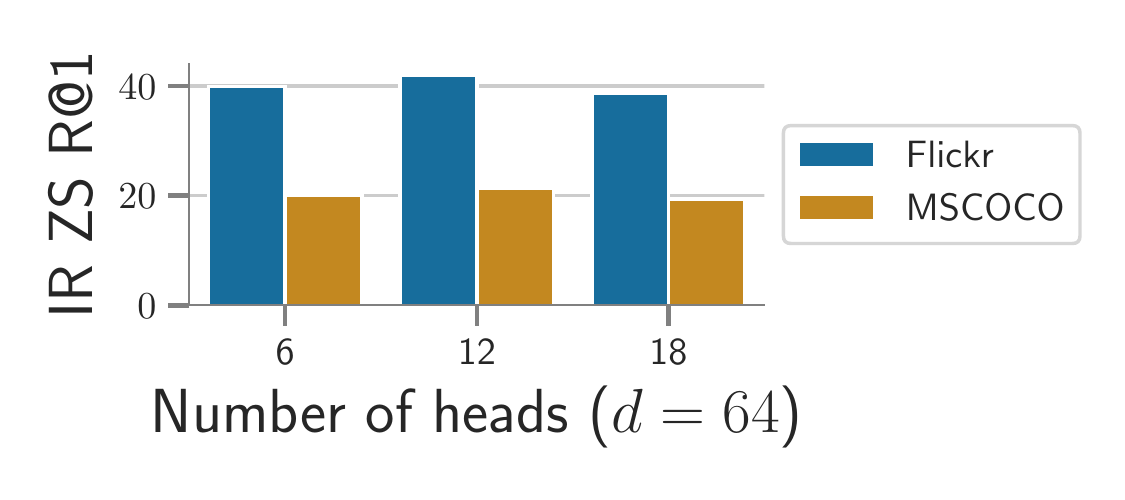} % first figure itself
                \caption{\small Heads: fixed dimension \label{fig:abl_heads_dims}}
        \end{subfigure}
        % \hfill

       \vspace{-0.5cm}
       % \vspace{-0.2cm}
\end{figure*}

\paragraph{Language-only Pretraining.}  Many multimodal transformers initialize language weights from a pretrained BERT model.
Similar to LXMERT, we find this hurts performance on our retrieval task; R@1 on \zsflickr decreases to $39.7$ and R@1 on \zsmscoco decreases to $20.4$.

\paragraph{Image-only Pretraining.} The object detector used to extract image features is another source of modality-specific pretraining.
% Another source of modality-specific pretraining in multimodal transformers is due to the image-region features taken from an object detector. 
%
We replace detection features with grid features taken from the last residual block of a ResNet-50 trained from scratch.\footnote{We fit images into a $384\times384$ square by resizing and padding to preserve the aspect ratio. As the total stride of ResNet-50 is $32$, a feature grid is of size $12\times12$, which we flatten to $144$ features and give as input along with the averaged features (for the \texttt{<CLS>} token) to our \modelname{}.}
Similarly to \citet{huang2020pixel}, this model is trained \emph{without} the MRM loss since  features aggregate information in the whole image, and as a result, masking specific regions is not straightforward. 
This model
performs slightly better than our base \modelname{}{} on \zsflickr ($43.4$ vs. $41.9$) and comparably on \zsmscoco ($21.3$ vs. $20.6$).
Though \citet{huang2020pixel} showed a detector can be replaced with an image classifier, we show that comparable results can be achieved without any image-only pretraining.

We conclude that careful consideration of pretraining datasets and their sampling methods is important in a model's performance -- the level of noise and the type of language in a dataset can be more significant than its size. 
Finally, the image-only and language-only pretraining are not crucial in training strong multimodal representations.

\subsection{Multimodal Attention}
\label{sec:attention_results}

We explore the impact of the number of attention heads and \coattn{} layers in our base multimodal transformer model 
before investigating the effect of different attention mechanisms. 
\paragraph{Number of Heads and Layers.}  
We test the importance of the number of heads in multi-head attention when fixing the total number of parameters by comparing models trained with one head, 3 heads, and 12 heads with query/key size of 768, 256, and 64, respectively. 
Increasing the number of heads to 12 leads to an improvement (Figure~\ref{fig:abl_heads_params}).
Next, we vary the number of heads (6, 12, and 18) but fix the query/key size to 64. 
We observe that increasing the number of heads up to 12 still leads to an improvement, but further increase results in poorer performance (see Figure~\ref{fig:abl_heads_dims}).

Consistent with \citet{lu2019vilbert}, increasing the number of layers (Fig.~\ref{fig:abl_layers}) helps up to a point, and then adding more layers degrades performance.

\begin{table}[!t]
\caption{\small{\modelname{} trained with \coattn{} (Co), \mergedattn (Merge), \langqattn{} (L-12 and L-24), \imageqattn{} (I-12 and I-24) (the number indicates the number of attention heads) and  \singleattn.}}
\resizebox{\linewidth}{!}{
\begin{tabular}{rccccccc}
\hline
\multicolumn{1}{c}{\multirow{2}{*}{R@1}} & \multirow{2}{*}{\textbf{Co}} & \multirow{2}{*}{\textbf{Merge}} & \multicolumn{4}{c}{\textbf{Asym. Attn.}}      & \textbf{Mod.}  \\
\multicolumn{1}{c}{}                     &                              &                                 & L-12 & I-12 & L-24 & I-24                      & \textbf{Spec.} \\ \hline
F. ZS                                & \textbf{41.9}                        & \multicolumn{1}{c|}{40.0}       & 24.4 & 31.3 & 33.6 & \multicolumn{1}{c|}{31.6} & 16.9           \\
F. FT                                &             \textbf{59.1}            & \multicolumn{1}{c|}{57.0}       & 45.1 & 48.4 & 52.5 & \multicolumn{1}{c|}{46.3} & 15.4           \\
M. ZS                                & \textbf{21.3}&  \multicolumn{1}{c|}{19.6}           & 13.8 & 16.1 & 17.0 & \multicolumn{1}{c|}{16.0} & 8.0            \\ \hline
\end{tabular}
}
\label{tab:attentiontypes}
 \vspace{-0.2cm}
\end{table}

\begin{figure*}
        \centering
        \caption{\small Comparing top-1 ranked images retrieved with models trained with the different attention mechanisms on the Flickr dataset. Correctly retrieved images are framed in green and the incorrect ones in red.
        \label{fig:qual-sm}}
        \vspace{-0.15cm}
        \includegraphics[width=\linewidth]{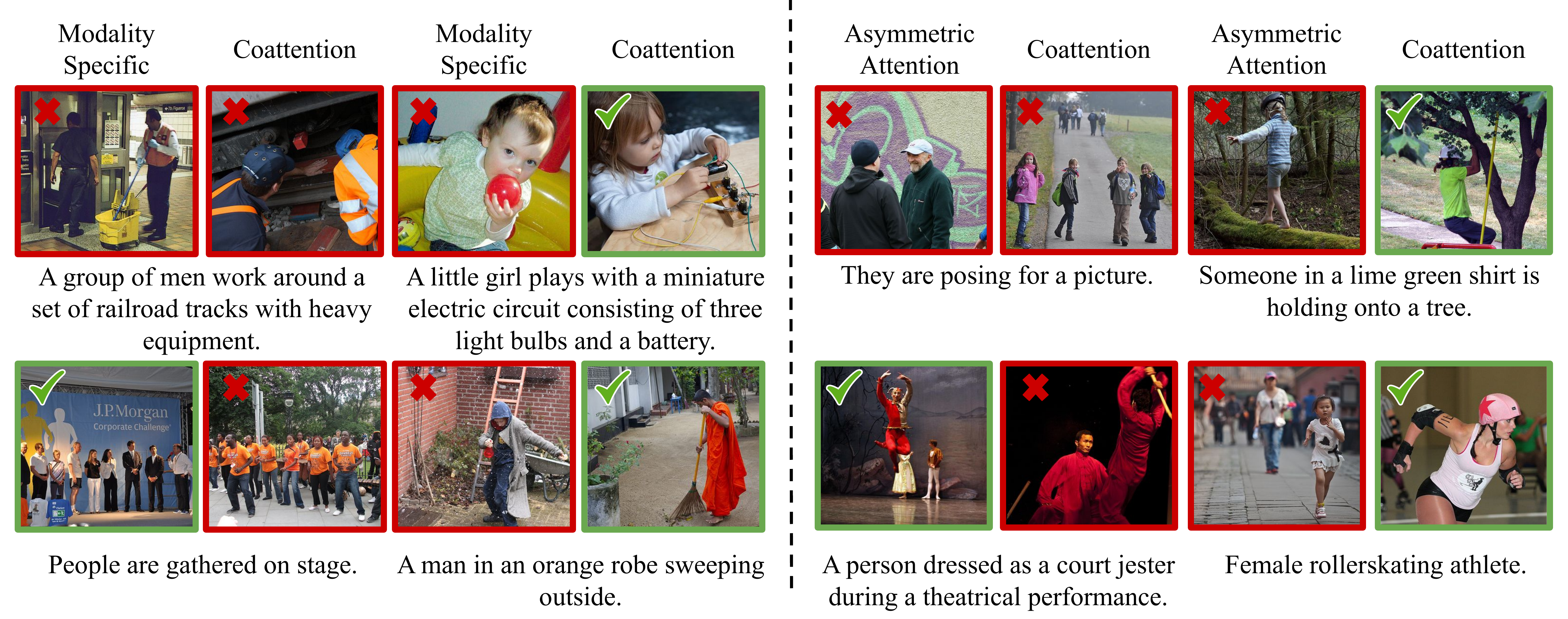} % second figure itself
        \vspace{-0.65cm}
\end{figure*}

\paragraph{Type of Attention Mechanism.}  
We perform an in-depth analysis on different types of attention explained in \Section{sec:mmattention} (see Table~\ref{tab:attentiontypes}).
We compare \coattn with \mergedattn -- these mechanisms both ``combine'' the image and language modalities; however, \coattn does so by taking keys/values and queries from opposite modalities, while \mergedattn{} shares keys and values across the modalities. 
When controlled for the number of parameters, \coattn performs marginally better than \mergedattn.
Both perform considerably better than asymmetric attention in which attention queries are over one modality.
  
The number of heads in an asymmetric attentions is half of the equivalent \coattn{}, so we experiment with asymmetric attention mechanisms with 12 heads (L-12, I-12) as well as 24 heads (L-24, I-24).
Increasing the number of attention heads for the asymmetric attention improves results, but the gap between our best-performing model with asymmetric attention (L-24) and \coattn{} is still quite large.

We also consider transformers with \singleattn{} where there is no cross-talk between the modalities through attention, but the model has the same number of parameters as our \modelname{} with \coattn{} and is trained with the same losses (\Table{tab:attentiontypes}, Mod. Spec. column). This model performs substantially worse than \modelname{}. % on both \zsflickr and \zsmscoco. 

To better demonstrate the strength of multimodal attention compared to asymmetric and modality-specific attention, we compare our models in \Table{tab:attentiontypes} to shallower and smaller models with \coattn{} on the \zsflickr task. Strikingly, our best-performing model \emph{without} multimodal attention with 24 attention heads and 12 layers (R@1 of 33.6; L-24 in \Table{tab:attentiontypes})  performs worse than the \coattn{} model with only one head (R@1 of  $38.2$; \Figure{fig:abl_heads_params}) or one multimodal layer (R@1 of $37.2$; \Figure{fig:abl_layers}).

Figures \ref{fig:qual-sm} % and \ref{fig:qual-asymmetric} 
shows example retrieval results comparing the asymmetric and modality specific attention to our coattention mechanism.
When the coattention mechanism retrieves the incorrect image, the image frequently includes important content from the sentence (\eg, in Figure \ref{fig:qual-sm} lower left, the image shows ``people gathered'', but they are not on stage).
Though other attention mechanisms retrieve images with some similarities to the text, the coattention mechanism retrieves fine details like ``lime green shirt'' and ``miniature electric circuit''.

A modality specific transformer model is computationally more efficient than models with multimodal attention because image and language features can be computed once and reused across image--text pairs; this means that single-modality transformers are faster for retrieval and thus would be more appealing in large-scale applications if their accuracy were higher.
We therefore investigate whether we can improve the single-modality transformer's poor performance by combining five \singleattn{} layers followed by one \coattn{} layer to introduce multimodal interaction. 
This model is as deep as our \modelname{}, but performs worse than our \modelname{} with one \coattn{} layer: R@1 of $33.1$ vs $37.2$ on \zsflickr and $16.7$ vs $19.0$ on \zsmscoco.

We conclude that \emph{multimodal attention} mechanisms, either \coattn{} or \mergedattn{}, are a key component to multimodal transformers' success. Moreover, a shallow or small model with multimodal attention outperforms deeper models with an inferior attention mechanism yet more parameters.
Finally, we show that a model's depth alone is not important; both multimodal attention and depth are needed for best performance.

\subsection{Losses}
\label{sec:loss_results}

We explore the degree to which MLM, MRM, and ITM losses contribute to our \modelname{} results. 
We then explore whether a contrastive formulation of the ITM loss -- used commonly in self-supervised representation learning and important for our baseline 
-- improves \modelname{}'s performance.

\paragraph{Comparing MLM, MRM, and ITM.} Table \ref{tab:loss1} shows performance of our models with different combinations of the masked modelling losses and the image-text loss.
With careful hyper-parameter tuning (in particular, decreasing the learning rate from $0.00176$ to $0.001$ and using cosine decay instead of polynomial decay) we can remove the MRM loss during pretraining and achieve comparable performance on our image retrieval tasks.
We found negligible difference when training our base \modelname{} with the different hyper-parameters. 
We note that our multimodal transformer trained on pixels (\Section{sec:data_results}) is also trained without a region modelling loss, yet performs similarly to our base \modelname{}.
Additionally, our finding is in line with the results of \citet{li2020oscar}, who achieve strong results without a region modelling loss.

\begin{table}[!t]
\caption{\small Zero-shot retrieval results (R@1) on models trained with different losses.}
% \vspace{-0.05cm}
\resizebox{\linewidth}{!}{
\begin{tabular}{lrr}
\toprule
       & Flickr-ZS & COCO-ZS \\ \midrule
       MRM + ITM & 20.2 & 9.7\\
       MLM + ITM & 41.1 & 22.4\\
       MRM + MLM + ITM & 41.9  & 21.3 \\
 \bottomrule
\end{tabular}}
\label{tab:loss1}
\vspace{-0.6cm}
\end{table}

\paragraph{Contrastive ITM Loss.}  
Contrastive losses (\eg,  \Equation{eq:itm-contrastive}) require sampling many negative examples to achieve good performance and thus can be computationally expensive \citep[\eg,][]{tian2019contrastive, miech20endtoend}. In models without multimodal attention (\eg, our baseline model), the computational cost is reduced by caching and reusing negative examples; in such models, since
image and text input are processed independently, once image and text features are calculated, they can be considered as negatives for all other training examples in the batch.
% On the other hand,  multimodal transformers, due to their multimodal attention, 
Due to their multimodal attention, multimodal transformers process image and text examples as pairs and thus cannot share image or text features across training examples. 
This limits the number of negatives available for these models to the maximum batch size that fits in memory. 
As a result, to study the role of a contrastive loss with a reasonable number of negatives, we consider our \modelname{} with one multimodal layer. 
We also examine whether a model with only \singleattn{} (here, we use 6 image and 12 language layers) benefits from a contrastive loss since it is easier to increase the negatives in a model without multimodal attention. 
In both models, we replace the image--text matching classification loss, \Equation{eq:itm-cls}, with a contrastive one, \Equation{eq:itm-contrastive}.

\Table{tab:loss2} compares the performance of a single-modality transformer trained with a classification loss to a model trained with a contrastive loss and 32 or 1024 negatives.
We observe a notable improvement with the contrastive loss and adding more negatives. 
We next compare the performance of our one-layer \modelname{} trained with a classification loss and a contrastive loss with 32 negatives (the max we could fit into memory). 
When training with the contrastive loss, we see no performance difference on \zsmscoco and a small performance degradation on \zsflickr.
This is surprising given the large body of research demonstrating the benefit of contrastive losses.
We conclude that the multimodal attention and MLM loss can help the model learn better representations without relying on stronger image--text losses.

\begin{table}[!t]
\caption{\small{R@1 with a classification ITM loss (cls) and contrastive ITM loss (con) for a \modelname{} with one multimodal layer (\modelname{}-1) and a model which only has modality specific attention (MSA).}}
\label{tab:loss2}
\resizebox{\linewidth}{!}{
\begin{tabular}{llrrr}
\toprule
      Model & Loss & Negatives & Flickr-ZS & COCO-ZS\\ \midrule
      MSA & Cls. & 1 &  15.0  & 6.9 \\
      MSA & Con. & 32 & 17.9  &  8.3  \\ 
      MSA & Con. & 1024 & 19.7  & 9.5  \\ \hline
      \modelname{}-1 & Cls. & 1 & 37.3   & 19.1\\
      \modelname{}-1 & Con. & 32 &  35.7 & 19.1 \\
 \bottomrule
\end{tabular}}
 \vspace{-0.3cm}
\end{table}

\section{Related Work}

Multimodal transformers are the first family of multimodal models to be pretrained on large data and applied to a range of different language and vision tasks \citep{lu2019vilbert, chen2019uniter, tan2019lxmert, li2020oscar, Li2020unicoder}. The recent image-text transformers share the same backbone but have slight differences in data preprocessing and other architectural choices. Notably, the UNITER model \citep{chen2019uniter} achieves state-of-the-art results on most existing image--language benchmarks by using a larger dataset and a number of different loss functions. \citet{huang2020pixel} removes the need for using image features (taken from a pretrained object detector) by training models on raw images (pixels). 
To combine image and text modalities, LXMERT \citep{tan2019lxmert} and \vilbert \citep{lu2019vilbert} propose coattention mechanisms, similar to the coattention originally proposed for VQA \citep{lu2016hierarchical}.
In \vilbert, feed-forward layers are applied after the coattention and self-attention layers, whereas in LXMERT, a feed-forward layer is only applied after the self-attention layer.

A few of our findings are similar to observations in prior work:
%the ablation studies of previous image--text transformers:
\textbf{(i)} LXMERT and \vilbert show that more layers improve results, %, at least up to a point.
\textbf{(ii)} \vilbert and UNITER show that more data boosts performance,
and \textbf{(iii)} LXMERT shows that transferring BERT weights is not beneficial.
In contrast to UNITER, we show that with the right hyper-parameters, the MRM loss is not needed.

Finally, while joint-space approaches to multimodal training are applied to multilingual data \citep{gella2017image,sigurdsson2020visual}, all existing multimodal transformers are applied to English; an interesting future direction is to extend these models to other languages.

\paragraph{Analyzing multimodal transformers.} 
Recent analysis work \citep{singh2020we, cao2020behind} has shed light on different aspects of multimodal transformer models. 
\citet{singh2020we} studies which pretraining data is best when fine-tuning two different multimodal transformer variants --\vilbert \citep{lu2019vilbert} and VisualBERT \citep{li2019visualbert}-- on four fine-tuned tasks, whereas we mainly focus on a zero-shot retrieval task across a variety of pretraining datasets, architectural choices, and loss functions.  
Our results are complementary to this work: \citet{singh2020we} observes that dataset size is not the only factor for good performance and pretraining datasets are better when they match the domain of a downstream task. 
We take a first step towards quantifying what it means for a pretraining dataset to be similar to a downstream task by analyzing the language used in the pretraining datasets and tasks (Section \ref{sec:data_results}).

\citet{cao2020behind} consider various probing % and visualization 
methods on two models (UNITER \citep{chen2019uniter} and LXMert \citep{tan2019lxmert}) to study what information is learned in pretrained models. \citet{cao2020behind} show that while representations become more similar in the last layers of models with merged attention, in coattention models, they are most similar at the first multimodal layer.  
They also observe that attention heads in merged attention models mostly focus on the language modality, only a few heads are specialized for cross-modality processing, and that attention heads are able to capture some image-text alignment.
Our comparisons of merged and coattention is performed in a more controlled setting than the work of \citet{cao2020behind} and \citet{singh2020we}: they compare two models trained by different researchers that include many small differences other than the attention mechanism; in contrast, we compare the attention mechanisms in the same modeling framework.

\section{Discussion}

We rigorously examined different aspects of training multimodal transformers (datasets, attention, and losses) that contribute to the quality of their learned representations. We focused on zero-shot image retrieval tasks to evaluate learned representations. Zero-shot tasks are advantageous because they directly measure what a model has learned and do not introduce confounds such as the size of a fine-tuning dataset and its experimental setup. 
At the same time, datasets do not always capture what they are designed to measure; \eg, \citet{akula2020words} show that models can do well on a referring expression task while ignoring the linguistic structure.
Thus, we argue that designing and curating specialized zero-shot evaluation tasks and datasets is an important future direction which will allow us to better understand our models' limitations.

We find the quality of language and the degree to which the language describes its corresponding image (noisiness) plays an important role in our results. Moreover, language-only and image-only pretraining do not notably contribute to the performance of multimodal transformers. These suggest curating less noisy image--text datasets to be more important than relying on single-modality datasets. Previous work has successfully removed some of the noise in automatically-harvested datasets through preprocessing \citep[\eg,][]{sharma2018conceptual} but such approaches are still limited in their robustness to noise, and the far from negligible degree of noise in large-scale real-world datasets \cite[\eg,][]{ordonez2011im2text,miech2019howto100m} still poses a challenge. An alternative approach is to aim to remove this noise by designing models that better tap into statistical regularities of image--text pairs \cite[\eg,][]{Duygulu:2002} and thus are more robust to noise.

We show that multimodal attention -- where each modality is informed by both modalities -- is crucial in these models' performance.
Smaller models with multimodal attention outperform deeper models with no or other multi-head attention mechanisms. 
This suggests that we can potentially train smaller models (than the existing multimodal transformers) for a given task, especially when the pretraining data is chosen carefully.
Moreover, with multimodal attention, we can achieve the best zero-shot retrieval results using a classification loss which uses only one negative example per image--text pair \citep[compare to a contrastive loss with 16384 negatives used in][]{tian2019contrastive} and also removes the need for mining more hard negatives \citep[]{faghri2017vse++}.

Additionally, we observe that comparable results can be achieved without the image (masked region modelling) loss in multimodal transformers. This suggests that our current models are not tapping into the useful signal in the image modality, presumably because of the image loss formulation.  An interesting future direction is designing better generative pretraining losses for images; previous work shows that the choice of loss 
significantly impacts the quality of language representations \citep{voita2020information}.

Finally, we believe that examining why and how multimodal transformers perform so well can guide future work in more effectively measuring progress in learning
rich visual-linguistic features.

\section*{Acknowledgements}
We would like to thank Angeliki Lazaridou, Andrew Zisserman, Phil Blunsom, Laura Rimell, and Stephen Clark for helpful feedback and conversations throughout the development of this work.  We would like to give special thanks to Aishwarya Agrawal for detailed comments and discussion on our initial paper draft.  Finally, we would like to thank Sebastian Borgeaud and Cyprien de Masson d'Autume for providing a language only BERT codebase.

% aishwarya, angeliki, az, phil, laura, steve, sebastian, cyprien

%Use unnumbered first level headings for the acknowledgments. All acknowledgments
%go at the end of the paper before the list of references. Moreover, you are required to declare 
%funding (financial activities supporting the submitted work) and competing interests (related financial activities outside the submitted work). 
%More information about this disclosure can be found at: \url{https://neurips.cc/Conferences/2020/PaperInformation/FundingDisclosure}.

%Do {\bf not} include this section in the anonymized submission, only in the final paper. You can use the \texttt{ack} environment provided in the %style file to autmoatically hide this section in the anonymized submission.

% \bibliographystyle{acl_natbib}
\interlinepenalty=10000

\end{document}